\documentclass[conference]{IEEEtran}
\usepackage{times}
\usepackage[numbers]{natbib}
\usepackage{multicol}
\usepackage[bookmarks=true]{hyperref}

\usepackage{cuted}
\hypersetup{
    colorlinks=true,
    linkcolor=blue,
    filecolor=magenta,      
    urlcolor=blue,
    citecolor=blue,
}
\usepackage{float}
\usepackage{color}
\usepackage{graphicx} 
\usepackage{tabularx}
\usepackage{amsmath}
\usepackage{xspace}
\usepackage{booktabs}
\usepackage{tcolorbox}
\usepackage{balance}

\usepackage{colortbl}
\usepackage{graphicx}
\let\labelindent\relax
\usepackage{enumitem}

\usepackage{wrapfig}

\usepackage{amsmath}
\usepackage{amsfonts}
\usepackage{bm}
\usepackage{vcell}
\usepackage{hyperref}
\usepackage{threeparttable}
\usepackage{scalerel}
\usepackage{float}
\usepackage{amssymb}
\usepackage{pifont}
\newcommand{\xmark}{\ding{55}}%
\usepackage{xcolor}
\newcommand{\edit}[1]{{#1}}

\newcommand{\method}{\textsc{Octopi}}
\newcommand{\dataset}{\textsc{PhysiCLeAR}}

\newcommand{\para}[1]{\vspace{0.4em}\noindent\textbf{#1}}

\usepackage[font=small,labelfont=sl,labelsep=period]{caption}

\usepackage{listings}
\usepackage{color}

\definecolor{dkgreen}{rgb}{0,0.6,0}
\definecolor{gray}{rgb}{0.5,0.5,0.5}
\definecolor{mauve}{rgb}{0.58,0,0.82}

\begin{document}
\title{\includegraphics[height=6mm]{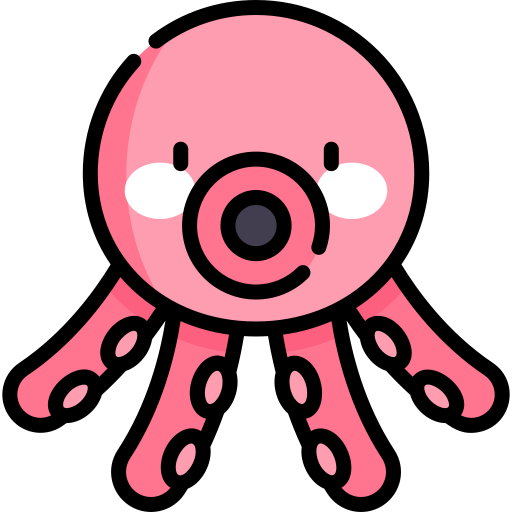}Octopi: Object Property Reasoning with Large Tactile-Language Models}

\author{\authorblockN{Samson Yu\authorrefmark{2}, Kelvin Lin\authorrefmark{2}, Anxing Xiao\authorrefmark{2}, Jiafei Duan\authorrefmark{4}, and Harold Soh\authorrefmark{2}\authorrefmark{3}
}
\authorblockA{\authorrefmark{2}Dept. of Computer Science, National University of Singapore}
\authorblockA{\authorrefmark{4}University of Washington, \authorrefmark{3}NUS Smart Systems Institute}
\authorblockA{Contact Authors: \texttt{samson.yu@u.nus.edu}, \texttt{harold@comp.nus.edu.sg}}
}

\maketitle
\begin{abstract}
Physical reasoning is important for effective robot manipulation. Recent work has investigated both vision and language modalities for physical reasoning; vision can reveal information about objects in the environment and language serves as an abstraction and communication medium for additional context. Although these works have demonstrated success on a variety of physical reasoning tasks, they are limited to physical properties that can be inferred from visual or language inputs. In this work, we investigate combining tactile perception with language, which enables embodied systems to obtain physical properties through interaction and apply commonsense reasoning. We contribute a new dataset \dataset, which comprises both physical/property reasoning tasks and annotated tactile videos obtained using a GelSight tactile sensor. We then introduce \method, a system that leverages both tactile representation learning and large vision-language models to predict and reason about tactile inputs with minimal language fine-tuning. Our evaluations on \dataset{} show that \method{} is able to effectively use intermediate physical property predictions to improve its performance on various tactile-related tasks. \dataset{} and \method{} are available at \href{https://github.com/clear-nus/octopi}{https://github.com/clear-nus/octopi}.
\end{abstract}


\IEEEpeerreviewmaketitle

\section{Introduction}
\label{sec:introduction}

For humans, touch is a crucial sense that provides physical information beyond what vision can provide (e.g., material properties, texture information, temperature), especially during occlusion. This in turn improves our ability to perform physical reasoning~\cite{melnik2023benchmarks, bakhtin2019phyre} and act in our world. Here, we are interested in enabling general purpose robots, specifically those empowered with large-language models (LLMs), to perform similar physical reasoning. 
While recent work has demonstrated that LLMs and large vision-language models (LVLMs) can provide an impressive level of commonsense and physical reasoning~\cite{team2023gemini, wu2023next, bai2023qwen, wei2023vary}, they are limited to either visual or text input modalities, and thus have limited performance in scenarios with visual ambiguity~\cite{gao2023physically}. 

\begin{figure}[t]
  \centering
  \includegraphics[width=\linewidth]{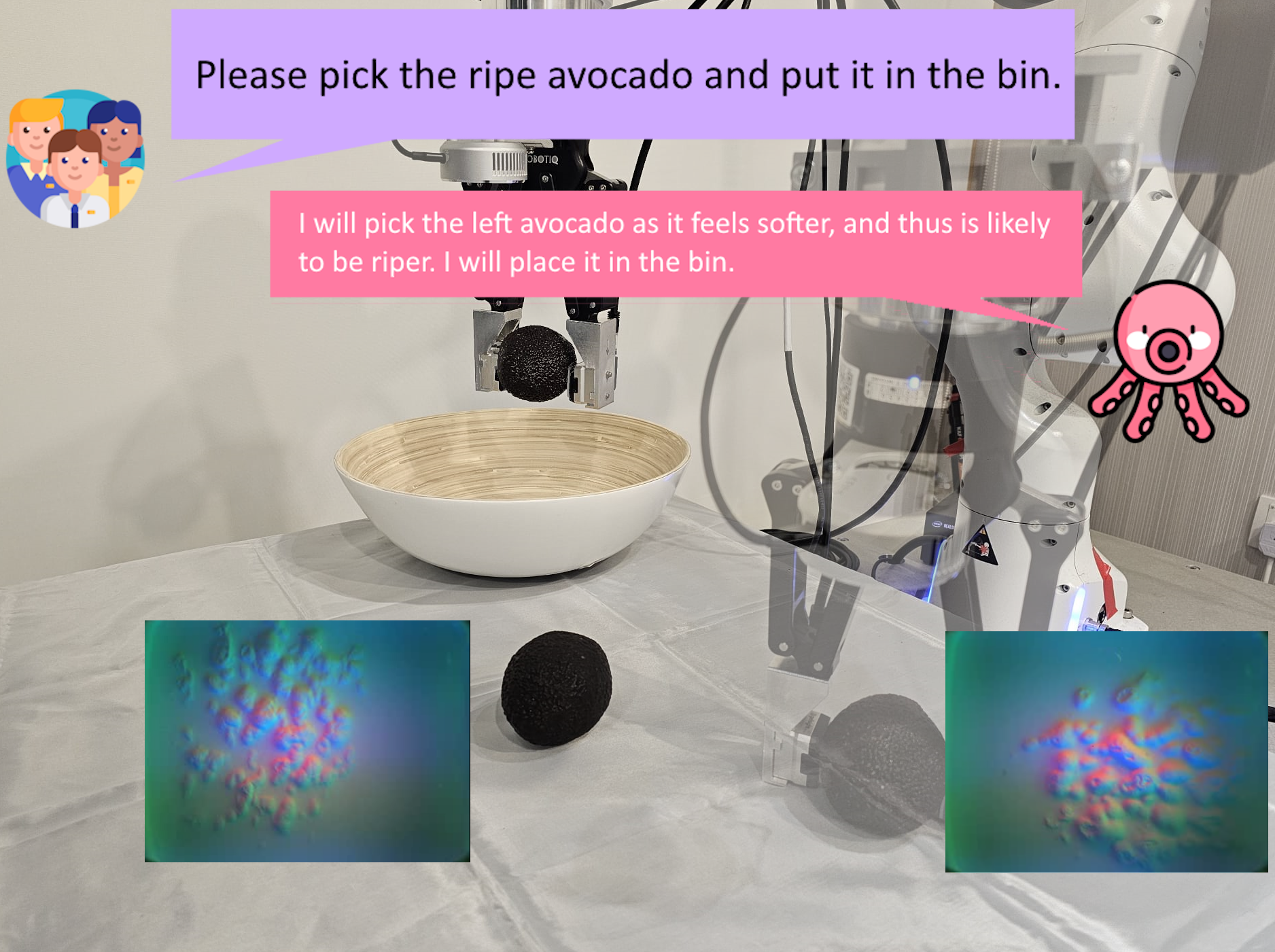}
  \caption{\edit{Avocado ripeness selection by combining tactile information with commonsense knowledge. Using inputs from its tactile sensor, \method{} identifies the left avocado as softer. Using commonsense reasoning, \method{} infers that it is ripe and fulfils the user's request.}}
  \label{fig:tasks}
  \vspace{0.5cm}
\end{figure}

This paper extends LVLMs to have the sense of touch. We posit that incorporating a tactile modality into LVLMs will enable better physical reasoning in real-world environments. 
\edit{As an example, Fig. \ref{fig:tasks} illustrates how commonsense knowledge is applied together with tactile information to complete a novel physical task. Here, the robot leverages its tactile inputs together with the LLM's commonsense knowledge (that ripe avocados are soft) to correctly select the ripe avocado.} We use visual-tactile sensors, i.e., the GelSight~\cite{s17122762}, which provides image frames that reveal physical object properties such as texture and hardness~\cite{7759057}.
However, there remains a significant domain gap between natural images that typical LVLMs are trained with and the tactile data. 

To bridge this gap, we contribute the \dataset{} dataset, which comprises GelSight images on a variety of real world objects, along with object labels and part annotations. \dataset{} complements existing tactile datasets~\cite{7759057,yuan2018active,gao2022objectfolder,yang2022touch,Gao_2023_CVPR} as it provides three physical property annotations, specifically hardness, roughness, and bumpiness, that have been used in prior research~\cite{takahashi2019deep, gao2016deep, purri2020teaching, article_chen, BERGMANNTIEST20102775, jiang2022robotic} and can be potentially inferred from the GelSight data. \dataset{} also includes an training and evaluation suite comprising five reasoning tasks, which can serve as a benchmark for the research community. 

Using \dataset, we develop \edit{\method{} (Object Comprehension with Tactile Observations for Physical Intelligence).}
\method{} is a LLaMA-based~\cite{touvron2023llama, touvron2023llama2} LVLM (Vicuna~\cite{vicuna2023}) equipped with a CLIP-based~\cite{radford2021learning} tactile encoder, whose representations have been aligned via projection. In experiments, we show that \method{} is \edit{able to use its} tactile modality to predict object properties and reason about scenarios including avocado ripeness.

\begin{figure*}[h]
    \centering
    \includegraphics[width=160mm]{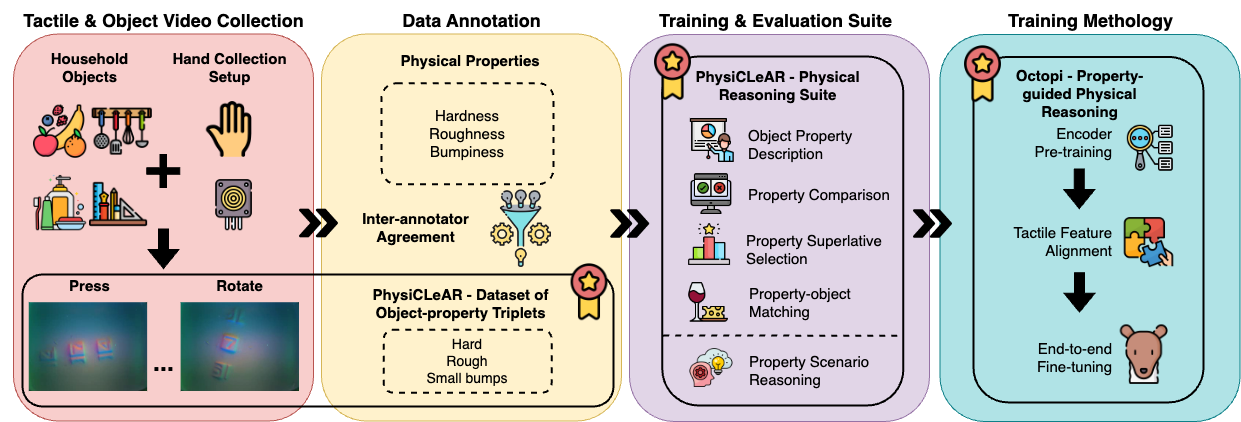}
    \caption{\textbf{\dataset{} and \method} (with key contributions starred). \edit{We collect tactile videos for everyday household objects by hand with two exploratory procedures: pressing and rotation. The videos are annotated by three annotators for three physical properties: hardness, roughness and bumpiness. \dataset{} leverages the videos and annotations for five language-driven physical description and understanding tasks. \method{} is a LVLM fine-tuned on \dataset{} for tactile-grounded physical understanding and reasoning.}}
    \label{fig:pipeline}
\end{figure*}

\begin{table*}
\centering
\caption{\textbf{Physical Property Details.} The physical object properties examined, along with their descriptions and semantic categories. \edit{}}
\begin{tabular}{l l l} 
 \hline
 \textbf{Property} & \textbf{Description} & \textbf{Categories} \\
 \hline
 Hardness & how easily an object's surface is deformed when pressed & soft, moderately hard, hard \\
 Roughness & how rough an object's surface is & smooth, slightly rough, rough \\
 Bumpiness & the size of bumps on an object's surface & no bumps, small bumps, big bumps \\
 \hline
\end{tabular}
\label{table:object_properties}
\end{table*}

\begin{table*}
\centering
\caption{\textbf{GelSight Dataset Comparisons.} \edit{\dataset{} provides physical property labels for tactile descriptions and physical reasoning across three physical properties. We further compare against existing datasets across three diversity measures. Property diversity refers to whether there are objects in the dataset that vary across the three properties we selected: hardness, roughness and bumpiness. Object diversity indicates whether there is more than one type of object in the dataset. Material diversity indicates the number of different materials in the dataset.}}
\begin{threeparttable}
\begin{tabular}{l | l | l l l}
 \hline
 \textbf{Dataset} & \edit{\textbf{Property Label Availability}} & \textbf{Property Diversity} & \textbf{Object Diversity} & \textbf{Material Diversity} \\
 \hline
 Hardness Dataset (2016) \cite{7759057} & \edit{Yes (only hardness)} & Yes & Yes & Medium \\
 Clothing Dataset (2018) \cite{yuan2018active} & \edit{Yes} & Yes & No (only clothing) & Low \\
 ObjectFolder 2.0 (2022) \cite{gao2022objectfolder} & \edit{No} & No (only hard objects) & Yes & Medium \\
 Touch and Go (2022) \cite{yang2022touch} & \edit{No} & Yes & Yes & High \\
 ObjectFolder-Real (2023) \cite{Gao_2023_CVPR} & \edit{No} & No (only hard objects) & Yes & Medium \\
 \hline
 \dataset & \edit{Yes} & Yes & Yes & Medium \\
 \hline
\end{tabular}
\end{threeparttable}
\label{table:datasets}
\end{table*}

\para{Contributions.} In summary, this paper makes the following key contributions:
\begin{itemize}
  \item A new GelSight dataset, \dataset, that exhibits property diversity, object diversity, and material diversity for selected physical properties.
  \item \method, a framework for physical reasoning that leverages vision-based tactile sensors and the commonsense reasoning capabilities of LLMs.
  \item An accompanying training and evaluation suite spanning five tasks and baseline results using \method.
\end{itemize}
We hope that \dataset{} and \method{} will spur research in tactile-enabled physical reasoning for embodied AI systems~\cite{duan2022survey}. 

\section{Related Work}
\label{sec:related-work}
In this section, we briefly review prior work on tactile representation learning with the GelSight sensor, large vision-language models (LVLMs) and language/vision-guided physical reasoning. \edit{There has been significant work in tactile-based manipulation and physical reasoning, and we refer readers desiring information on these topics to relevant survey papers~\cite{melnik2023benchmarks, duan2022survey, zhu2020dark, manipulationarticle, doi:10.1177/027836490001900702}.}

\para{Tactile Representation Learning with GelSight.}
Tactile representation learning has advanced significantly in recent years as robotic manipulation often requires more precision beyond what can be provided by vision alone~\cite{taunyazov2020event}. Among the available tactile sensors, vision-based sensors have gained popularity due to their high-resolution image outputs and versatility. In particular, the GelSight sensor has been used in recent work~\cite{6619008, 7759057, s17122762, huang2022understanding, yuan2018active} for inferring physical properties (e.g. hardness, texture and liquid volume) and to manipulate objects~\cite{tian2019manipulation}. A key benefit of the GelSight is that its image outputs can be easily processed by modern deep learning methods~\cite{huang2022understanding}. As a result, popular vision algorithms have been used for tactile representation learning with GelSight~\cite{zandonati2023investigating, cao2023learn}.  In our work, we exploit recent advances in tactile representation learning to extend the capabilities of LVLMs to reason about vision-based tactile input.

\para{Large Vision-Language Models.}
Recent advancements in LLMs have spurred a significant increase in efforts to integrate vision models with LLMs, exemplified by Flamingo~\cite{alayrac2022flamingo}, BLIP-2~\cite{li2023blip}, and MiniGPT-v2~\cite{chen2023minigpt}. These Large Vision-Language Models (LVLMs) have shown remarkable effectiveness in utilizing web-scale image-text data for image-based reasoning, benefiting a range of applications from robotics~\cite{brohan2023rt,duan2023ar2} to medical imaging~\cite{saygin2023quilt}. Very recent work involves developing LVLMs that can process video content~\cite{li2023videochat,maaz2023video}, enabling reasoning over dynamic visual information, or integrate multi-modal sensory data \cite{zhang2023meta}.

\para{Physical Reasoning with Language and Vision as Context.} 
The exploration of physical reasoning in conjunction with language predates the emergence of LLMs. Early studies focused on assessing model proficiency in physical reasoning. For example, the PIQA~\cite{bisk2020piqa} benchmark evaluates models on physical common sense, whereas PROST\cite{aroca2021prost} examines their understanding of physical reasoning concepts. Subsequent advancements in language grounding have led to works such CLEVRER~\cite{yi2019clevrer}, PIP~\cite{duan2022pip}, SPACE~\cite{duan2021space} and Phys101~\cite{wu2016physics}, which investigate the acquisition of physical reasoning skills from visual inputs. 

In the emerging LLM era, research has focused on object-centric physical reasoning in LLMs. This involves evaluating various LLMs for their physical reasoning capabilities, e.g., NEWTON~\cite{wang2023newton}, and employing Vision-Language Models (VLMs) to predict physical properties that are then used to facilitate reasoning, as demonstrated in physically-grounded VLMs \cite{gao2023physically}. Unlike previous studies that primarily address physical reasoning through the integration of vision and language, \method{} stands out as the one of the first models capable of processing tactile images alongside language instructions to enable physical reasoning. \edit{There has been very recent work~\cite{hong2024multiply} that uses simulated tactile inputs with LLMs, but we focus on real tactile data.} Concurrent work~\cite{fu2024touch,yang2024binding} also explores real-world tactile data but our work features physical property annotations and a test suite comprising scenario reasoning tasks, and experiments using \method{} to evaluate the utility of physical property inference.

\section{\dataset{} - Tactile and  Physical Understanding Training \& Evaluation Suite}
\label{sec:Feelang}

This section describes \dataset, which comprises a tactile dataset with physical property and object-part annotations, along with a training and evaluation suite. 

\subsection{Physical Property Selection}
In this work, we focus on three object properties: \textit{hardness}, \textit{roughness} and \textit{bumpiness}. We list each property's description and categories in Table \ref{table:object_properties}. Briefly, \textit{hardness} is characterized by the extent of surface deformation when subjected to pressure; \emph{roughness} pertains to the texture of the surface; and \emph{bumpiness} describes the prominence of surface protrusions. The \emph{hardness} of an object correlates with its compliance and thermal characteristics. In contrast, \emph{roughness} and \emph{bumpiness} are attributes influenced by the surface's friction coefficient~\cite{article_chen}.

The selection of \emph{hardness}, \emph{roughness}, and \emph{bumpiness} as physical attributes in our research is grounded in their relevance for physical reasoning~\cite{takahashi2019deep, gao2016deep, purri2020teaching, article_chen, BERGMANNTIEST20102775, jiang2022robotic}. Generally, static physical properties of objects are categorized into geometric (e.g., size), material (e.g., hardness), and affective (e.g., comfort)~\cite{Sakamoto2017ExploringTP}. Our study predominantly addresses material properties, as we deemed geometric and affective properties too challenging to ascertain using the GelSight. The choice of these specific properties was also informed by the data collection methodology~\cite{klatzky2016haptic}, tailored to the limitations and strengths of the GelSight sensor, including considerations for its sensitivity and durability.

\subsection{Dataset Collection \& Annotation}
\label{sec:dataset}

To facilitate the grounding of our physical reasoning on tactile inputs, we collected a dataset of 74 everyday objects, totalling 408 tactile videos \edit{and corresponding videos showing the object as the data was collected}. These objects were selected to span across our three selected properties, with variations across object types and materials. Detailed comparisons between \dataset{} and existing GelSight datasets can be found in Table \ref{table:datasets}. 

\edit{
The GelSight data was collected by-hand to mitigate risk of damaging the sensors and due to the challenge of securing different parts of irregularly-shaped objects while performing the required sampling motions.}
For each selected object, we captured up to seven tactile videos for each distinct region identified by a human evaluator. This process involved a two-step procedure: initially pressing the GelSight sensor against the object to capture pressure readings, followed by rotating the sensor to acquire shear readings. Each video generated from a single GelSight sensor reading constitutes an individual data point within our dataset.

Annotations of the physical properties were carried out by three independent annotators, with the average score used as the final annotation for each data point. \edit{Annotators were provided with both the tactile videos and the objects. Each property has three categories, and annotators were given the following guidelines for labeling each property:
\begin{itemize}
\item \textbf{Hardness:} The label \textit{soft} is  for objects that are compressible with little force, \textit{moderately hard} for objects that are compressible with moderate force, and \textit{hard} for objects that are incompressible even with a large pressing force.
\item \textbf{Roughness:} \textit{smooth} is for objects that present very minimal or no resistance when we slide our finger across its surface, \textit{slightly rough} for objects with slight resistance, and \textit{rough} for objects with significant resistance.
\item \textbf{Bumpiness:} \textit{no bumps} is for objects with no visible protrusions on its surface, \textit{small bumps} for objects with protrusions smaller than $\approx 1/4$ of the tactile image upon contact, and \textit{big bumps} for objects with protrusions larger than $1/4$.
\end{itemize}
This process yielded over 1,200 annotations and we observed high inter-annotator agreement scores (ICC3k of 0.894 (hardness), 0.979  (roughness), and 0.792 (bumpiness)). For reference, a score above 0.75 is considered good or excellent reliability.} The dataset was subsequently divided into three distinct subsets (training, validation, and testing) following an 80-10-10 split. This division resulted in 60 objects for training and 7 objects each for validation and testing. 

\subsection{Training \& Evaluation Suite}
\label{sec:suite}

\begin{table*}
\centering
\caption{\textbf{\dataset{} Task Details.} \edit{\dataset{} contains five physical description and understanding tasks. We give each task's motivation and indicate whether they are used for \method's training and/or evaluation.}}
\begin{tabular}{l l c c}
 \hline
 \textbf{Task} & \textbf{Motivation} & \textbf{Training [T]} & \textbf{Evaluation [E]} \\
 \hline
 Object Property Description (OPD) & LLM learns to describe tactile videos using physical properties. & \checkmark & \checkmark \\
 Property Comparison (PC) & Align LLM physical understanding with physical property descriptions. & \checkmark & \checkmark  \\
 Property Superlative Selection (PSS) & Align LLM physical understanding with physical property descriptions. & \checkmark & \checkmark  \\
 Property-object Matching (POM) & Align LLM object understanding with physical property descriptions. & \checkmark & \checkmark \\
 Property Scenario Reasoning (PSR) & Evaluate LLM's ability to resolve scenarios using physical properties. & \xmark & \checkmark \\
 \hline
\end{tabular}
\label{table:task_details}
\end{table*}

\begin{table*}
\centering
\caption{\edit{\textbf{\dataset{} Task Prompt Setup.} We provide the language prompts for each task in \dataset. The LLM (i.e. ASSISTANT) responses for each task are ground-truth tactile descriptions, except the last response in italics for PSR. This italicized response is generated by \method-13b, since PSR is not used for training and hence there is no ground-truth response.}}

\begin{tabular}{l l}
 \hline
 \textbf{Task} & \textbf{Prompt Setup} \\
 \hline
 OPD & \textbf{USER}: Describe the physical properties of \scaleleftright{<}{tact\_start}{>}$T_{1},...,T_{N}$ \scaleleftright{<}{tact\_end}{>}. \\
 & \textbf{ASSISTANT}: The surface feels somewhat hard, lacks noticeable roughness and there are no textural variations or fine ridges. Overall, it \\
 & presents a moderately hard and smooth surface with no bumps. \\
 \hline
 
 PC & \textbf{USER}: Is the object in the tactile video \scaleleftright{<}{tact\_start}{>}$T_{1,1},...,T_{1,N}$\scaleleftright{<}{tact\_end}{>} covered with bigger bumps than the one in \scaleleftright{<}{tact\_start}{>} \\
 & $T_{2,1},...,T_{2,N}$\scaleleftright{<}{tact\_end}{>}? Describe both objects before answering. \\
 & \textbf{ASSISTANT}: First object: The surface feels firm and somewhat flexible, is smooth and there are big bumps and ridges. Overall, it presents a \\
 & hard and smooth surface with big bumps. Second object: The surface feels soft, exhibits no roughness and there are small imperfections and \\
 & ridges. Overall, it presents a moderately hard and slightly rough surface with small bumps. Conclusion: Yes, the first object is covered with \\
 & bigger bumps. \\
 \hline
 
 PSS & \textbf{USER}: Given three tactile videos: a) \scaleleftright{<}{tact\_start}{>}$T_{1,1},...,T_{1,N}$\scaleleftright{<}{tact\_end}{>}, b) \scaleleftright{<}{tact\_start}{>}$T_{2,1},...,T_{2,N}$\scaleleftright{<}{tact\_end}{>}, c) \scaleleftright{<}{tact\_start}{>} \\
 & $T_{3,1},...,T_{3,N}$\scaleleftright{<}{tact\_end}{>}. Describe each object and then select the smoothest. \\
 & \textbf{ASSISTANT}: a) The surface feels yielding, exhibits significant roughness and there are small bumps and ridges present. Overall, it presents \\
 & a soft and rough surface with small bumps. b) The surface feels somewhat firm and flexible, exhibits no roughness and there are big bumps on \\
 & the surface. Overall, it presents a moderately hard and smooth surface with big bumps. c) The surface feels soft and yielding, exhibits a slight \\
 & roughness and there are tiny textural variations due to the fibrous structure. Overall, it presents a soft and slightly rough surface with small \\
 & bumps. Conclusion: b) is the smoothest. \\
 \hline
 
 POM & \textbf{USER}: Given three tactile videos: a) \scaleleftright{<}{tact\_start}{>}$T_{1,1},...,T_{1,N}$ \\
 & \scaleleftright{<}{tact\_end}{>}, b) \scaleleftright{<}{tact\_start} {>}$T_{2,1},...,T_{2,N}$\scaleleftright{<}{tact\_end}{>}, c) \scaleleftright{<}{tact\_start}{>}$T_{3,1},...,T_{3,N}$\scaleleftright{<}{tact\_end}{>}. Describe the object in each video, then \\ 
 & match each video to one of the following objects in alphabetical order: 1) a mandarin orange, 2) a cotton ball, 3) a roll of masking tape. \\
 & \textbf{ASSISTANT}: a) The surface feels firm, exhibits a slight roughness and there are small textural irregularities and fine ridges. Overall, it \\ 
 & presents a hard and slightly rough surface with small bumps. b) The surface feels soft and yielding, exhibits no roughness and there are no \\ 
 & prominent bumps or ridges. Overall, it presents a soft and smooth surface with no bumps. c) The surface feels soft, exhibits a slight roughness \\ 
 & and there are tiny bumps present. Overall, it presents a soft and slightly rough surface with small bumps. Conclusion: a) is a roll of masking \\ 
 & tape, b) is a cotton ball and c) is a mandarin orange. \\
 \hline
 
 PSR & \textbf{USER}: Describe these two tactile videos: a) \scaleleftright{<}{tact\_start}{>}$T_{1,1},...,T_{1,N}$\scaleleftright{<}{tact\_end}{>}, b) \scaleleftright{<}{tact\_start} {>}$T_{2,1},...,T_{2,N}$\scaleleftright{<}{tact\_end}{>}. \\
 & \textbf{ASSISTANT}: a) The surface feels hard and solid, exhibits a lack of roughness and there are no textural variations or fine ridges. Overall, it \\ 
 & presents a hard and smooth surface with no bumps. b) The surface feels yielding, has a subtle roughness and there are small irregularities and \\
 & bumps. Overall, it presents a soft and slightly rough surface with small bumps. \\
 & \textbf{USER}: In an emergency, which of the objects above can be used to break through thin ice covering a car outside? Select only one most \\
 & appropriate object for this scenario based on physical property descriptions of the two objects. Use the format `The most suitable object is x), \\
 & because xxx'. \\
 & \textbf{ASSISTANT}: \emph{The most suitable object is a) a hard and smooth surface with no bumps, because it is more likely to penetrate the thin ice} \\
 & \emph{covering a car exterior and not get stuck.} \\
 \hline
\end{tabular}
\label{table:task_prompts}
\end{table*}

\dataset's training and evaluation suite comprises five physical reasoning tasks (Table \ref{table:task_details}). All five tasks use tactile data and natural language instructions as inputs (Table \ref{table:task_prompts}). Since the tactile data is in video form, we follow prior LVLM work and represent it as a sequence of frames: $X_{1},...,X_{N}$. We further detail each task's motivation, setup, evaluation details \edit{and whether they are used for training [T] and/or evaluation [E]} below:

\para{Object Property Description (OPD) [T, E].} 
This task addresses property-based description: generating both unstructured and structured descriptions of an object's hardness, roughness, and bumpiness from tactile videos. It parallels existing image or video captioning tasks, which use natural visuals, and aids the LLM in interpreting tactile signals. The unstructured description provides more complex descriptions in cases where they are visible in the tactile video (e.g. ``fibrous structure" for toilet paper or ``grains" for a scoop of rice). The language instructions are variants of ``Describe the physical properties of \scaleleftright{<}{tact\_start}{>}$T_{1},...,T_{N}$\scaleleftright{<}{tact\_end}{>}." The unstructured description is generated using ChatGPT 3.5 and manually cleaned to produce a diverse description of our objects based on our three physical properties. The structured description of an object's physical properties using our annotations is formatted as: ``Overall, it presents a \textit{hardness\_label} and \textit{roughness\_label} surface with \textit{bumpiness\_label}."

\para{Property Comparison (PC) [T, E].} Given two tactile videos, each of a different object, a specified physical property, and its comparative adjective, determine whether the comparative adjective accurately describes the two videos. 
From a training perspective, this task helps a model distinguish between the various descriptions of physical properties, thereby aligning its comprehension of physical characteristics with our defined categories of \emph{hardness}, \emph{roughness}, and \emph{bumpiness}. This alignment ability may improve a model's ability to interpret and reason about the physical world in a manner consistent with human understanding. 

\para{Property Superlative Selection (PSS) [T, E].} For three tactile videos, each of a different object, and a specified physical property and its superlative adjective (e.g. hardest for the hardness property), choose the video that the superlative adjective best describes. This task is similar to the PC task and helps the LLM align its physical understanding with that of our physical property descriptions. Furthermore, since prior work has shown that LLMs might perform differently when the polarity of the comparative adjective changes \cite{wang2023newton}, this task seeks to enhance the LLM's resilience to various comparative descriptions of physical properties.

\para{Property-object Matching (POM) [T, E].} This task requires matching physical properties to objects: given three tactile videos (each featuring a different object) and three specified objects, the goal is to correctly associate each video with an object. This helps to align a model's existing knowledge of object properties with our haptic perception, as our annotations are based on human touch and serve as the reference for the physical properties and their labels.

\para{Property Scenario Reasoning (PSR) [E].} 
We provide two tactile videos, each showcasing a different object, along with a real-world scenario that relies on one or more of our defined physical properties. The task is to choose the video that represents the object whose physical properties best meet the scenario's demands. This approach allows us to assess a model's physical reasoning capabilities. Details of the scenarios are presented in Table \ref{table:scenario}.

\begin{table*}
\centering
\caption{\textbf{Scenario Questions with Selected Object and/or Properties.} \edit{We list the three scenarios we use to evaluate our model's physical reasoning capabilities. For each scenario, we either specify the target object (along with the properties it has) or the target properties to select the target and non-target objects.}}
\begin{tabular}{l l l} 
 \hline
 \textbf{Scenario Question} & \textbf{Target Object} & \textbf{Target Properties} \\
 \hline
 Which object is most suitable for removing stains from a non-stick pan without scratching it? & Microfiber cloth & hardness, roughness \\
 Which object would be most easily grippable when wet and slippery? & - & roughness, bumpiness \\
 In an emergency, which of the objects above can be used to break through thin ice covering a car outside? & - & hardness \\
 \hline
\end{tabular}
\label{table:scenario}
\end{table*}

\section{\method{} - Vision-language Property-guided Physical Reasoning}
\label{sec:octopi}

\begin{figure}
    \centering
    \includegraphics[width=\linewidth]{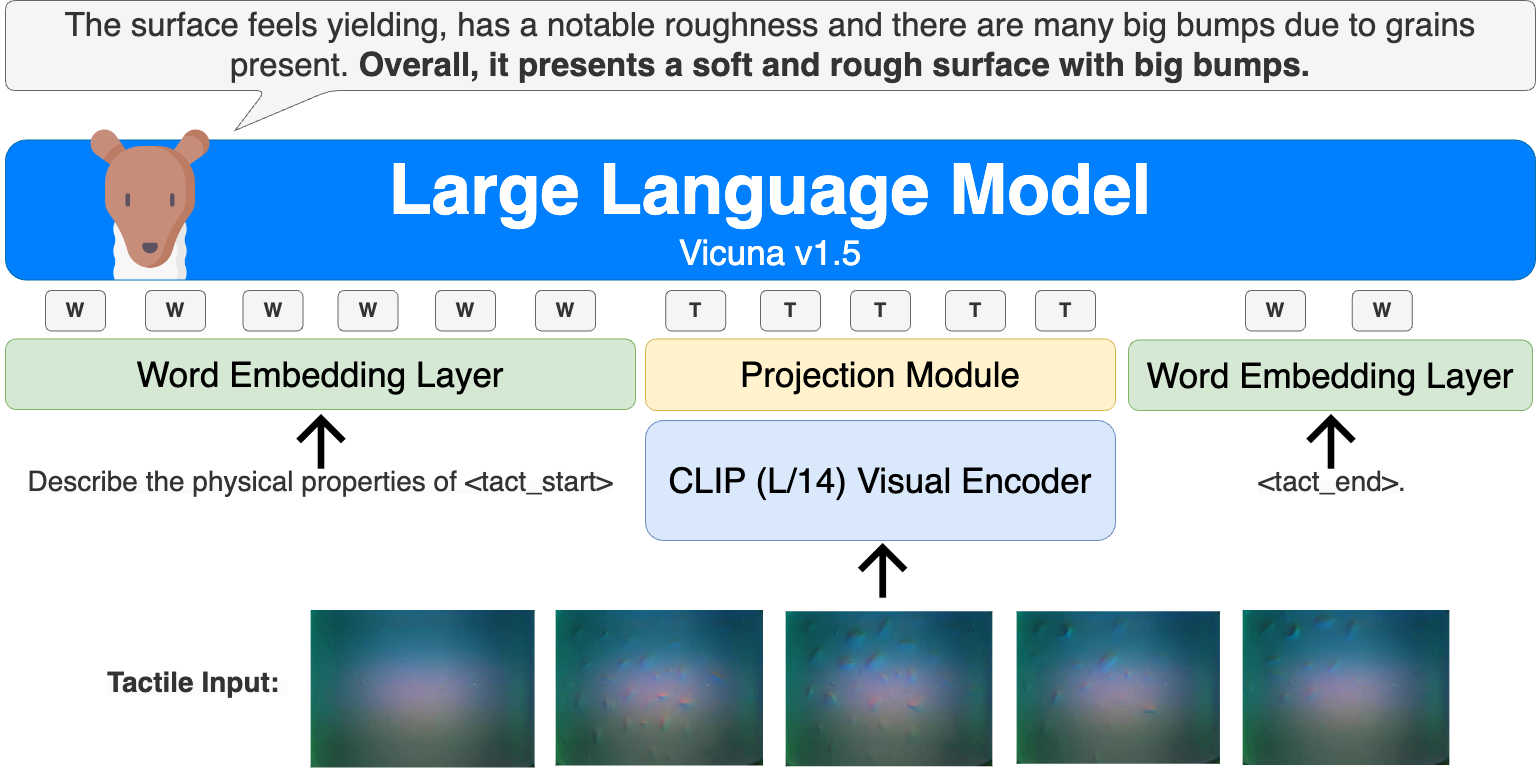}
    \caption{\edit{\textbf{\method{} Framework.} Our framework consists of CLIP's visual encoder, a projection module with two linear layers, and Vicuna v1.5 as the LLM. Language embeddings are derived through tokenization and then Vicuna's word embedding layer, with \scaleleftright{<}{tact\_start}{>} and \scaleleftright{<}{tact\_end}{>} being newly trained word embeddings indicating the start and end of a tactile frame sequence from a single tactile sensor. Tactile frames are fed into the visual encoder followed by the projection module to derive tactile embeddings with the same dimension as the word embeddings.}}
    \label{fig:model}
\end{figure}

The \method{} framework comprises three trained components: 1) tactile input encoder, 2) projection module, and 3) LLM, similar to prior LVLM models~\cite{liu2023visual, maaz2023video, zhang2023video}. A summary of our overall framework is shown in Fig. \ref{fig:model}.

We leverage the capabilities of pre-trained vision models, notably the CLIP \cite{radford2021learning} visual encoder ViT-L/14, as the foundation for our tactile encoder to derive meaningful feature representations. The encoder’s output is then mapped to the LLM's word embedding space using a projection module, typically consisting of one or two trainable layers. Our projection module, inspired by LLaVA \cite{liu2023visual, liu2023improved}, employs two linear layers with an intermediate GELU activation \cite{hendrycks2016gaussian}. Lastly, the LLM serves as the language understanding component in \method{}. The performance of the LLM is largely influenced by its pre-training datasets. We utilize the open-source LLaMA-based LLM, Vicuna \cite{vicuna2023}, recognized for its dialogue capabilities.

The inference process is illustrated in Fig. \ref{fig:model}. \method{} receives an instruction to evaluate the physical properties of uncooked rice. The text is tokenized and fed into the LLM's word embedding layer to produce word \textit{[W]} embeddings. A sequence of five tactile images is processed through the tactile encoder, with the output embeddings sent to the projection module to obtain the final tactile \textit{[T]} embeddings. Newly trained word embeddings, represented by \scaleleftright{<}{tact\_start}{>} and \scaleleftright{<}{tact\_end}{>}, mark the beginning and end of the tactile data, respectively. These tactile embeddings are then merged with the word embeddings at designated positions to form the final instruction embeddings for the LLM.

We follow a three-step training methodology: (i) encoder fine-tuning, (ii) tactile feature alignment, and (iii) end-to-end fine-tuning. In the following, we describe each of these steps in greater detail.

\subsection{Encoder Fine-tuning}
\label{sec:encoder_fine_tuning}
Existing LVLM models take natural videos as input and can use CLIP's visual encoder without modification. However, our work involves vision-based tactile inputs, which marks a significant distribution shift from natural images, necessitating additional fine-tuning to derive useful representations from these inputs.

We fine-tune our visual encoder to obtain useful representations from tactile inputs using multitask physical property classification. We adopt the architecture of ViFi-CLIP~\cite{rasheed2023fine} so that our visual encoder can be trained on video inputs. In ViFi-CLIP, frame-level embeddings from CLIP's visual encoder are average-pooled to obtain a video-level representation. 

We then append learnable prompts to the pre-trained CLIP visual encoder ViT-L/14 following Visual Prompt Tuning (VPT) \cite{jia2022visual} and initialize ViFi-CLIP's visual encoder with the new visual encoder. Specifically, we attach 8 task-specific learnable prompts and a shared linear layer to the input sequence of each Transformer \cite{vaswani2017attention} layer in the visual encoder and freeze the entire pre-trained Transformer backbone.

Finally, we add three separate classification heads to ViFi-CLIP, each of which predicts a label for one property (i.e. hardness, roughness or bumpiness), and train all three classification heads simultaneously using the cross-entropy loss. The model achieving the highest combined validation accuracy --- correctly predicting all three properties for an object --- is selected.

\subsection{Tactile Feature Alignment}

\edit{We discard the fine-tuned CLIP's classification layers and use the outputs from its visual encoder as output embeddings. To align the output embeddings from the fine-tuned visual encoder with the LLM, the  projection module is trained on language annotations while the encoder and the LLM are frozen. We also fine-tune the embedding layer  due to the two new word tokens (i.e. \scaleleftright{<}{tact\_start}{>} and \scaleleftright{<}{tact\_end}{>}).}

\subsection{End-to-end Fine-tuning}
\label{sec:end_to_end_fine_tuning}

Finally, we used end-to-end fine-tuning to improve the coherence of the LLM's responses and increase the similarity between its responses and the language annotations. In this stage, only the visual encoder is frozen while the word embedding layer, projection module, and LLM are fine-tuned. We fine-tune the LLM using low-rank adaptation (LoRA) \cite{hu2021lora} for parameter-efficient fine-tuning.

\section{Experimental Setup}
\label{sec:experimental_setup}
In this section, we evaluate the physical property prediction and reasoning capabilities of our proposed method. We design several experiments to answer the following questions:
\begin{enumerate}
    \item Are our physical property predictions useful for \method{} to reason about everyday scenarios?
    \item Can \method{} be used in real robots to help them accomplish tasks using physical reasoning?
    \item Can \method's understanding of the physical properties generalize to the unseen daily life objects?
\end{enumerate}

\subsection{Data Processing}
\label{sec:data_processing}
The tactile videos were processed into frames. To focus on a few salient frames for better efficiency, we selected frames that have the top 30\% total pixel intensity difference with their preceding frames. We randomly sampled 5 frames \edit{from these salient frames} during training and selected 5 frames at uniform intervals from the first salient frame during evaluation. Data augmentation was performed during training in the form of random horizontal and vertical flips with 50\% chance for each flip.

\subsection{Training Hyperparameters}
Encoder fine-tuning was performed for 30 epochs using the AdamW optimizer \cite{loshchilov2017decoupled} with no weight decay, a learning rate of $10^{-3}$, batch size of 32, and a cosine annealing learning rate schedule. During tactile feature alignment, the projection module is trained using 8k \dataset{} samples using the AdamW optimizer~\cite{loshchilov2017decoupled} with no weight decay, a learning rate of $2\times 10^{-5}$, batch size of 16, and a cosine annealing learning rate schedule.

For end-to-end fine-tuning, both the projection module and the LLM's LoRA parameters are trained using 3k \dataset{} samples using the AdamW optimizer \cite{loshchilov2017decoupled} with no weight decay, batch size of 16 and a cosine annealing learning rate schedule. Learning rates of $2\times 10^{-5}$ and $2\times 10^{-4}$ were used for the projection module and the LLM, respectively. We use a scaling factor of 256, a rank of 128, and a dropout rate of 0.05 for LoRA.

\edit{
\subsection{Training Requirements}
\label{sec:requirements}
Encoder fine-tuning took 6 hours and required less than 5GB of GPU VRAM. Tactile feature alignment together with end-to-end fine-tuning took 5 hours for \method-7b and 6.5 hours for \method-13b. We used 1 NVIDIA RTX A6000 for \method-7b and 2 NVIDIA RTX A6000s for \method-13b.
}

\section{Experimental Results}
\label{sec:results}
To address the above questions, we evaluated \method{} using (i) accuracy on the physical understanding tasks in \dataset's test set, (ii) accuracy on scenario reasoning tasks, (iii) task success rate on a real robot, and (iv) property prediction accuracy on unseen objects. \edit{We tested two versions of \method, \method-7b and \method-13b, which use Vicuna-7b v1.5 and Vicuna-13b v1.5 as their LLMs respectively.}

\subsection{Tactile-grounded Physical Understanding with Object Property Descriptions}

\begin{table}
\centering
\caption{\textbf{Results on \dataset{} Physical Understanding Tasks.} \edit{\method's performance on physical understanding tasks improves with object property descriptions (OPD). Performance also increases with larger LLM size, with \method-13b outperforming \method-7b across all three tasks.}}
\begin{tabular}{l c c c}
 \hline
 & \textbf{Random} & \textbf{\method-7b} & \textbf{\method-7b (no OPD)} \\
 \hline
  PC & 33.33 & \edit{\textbf{48.10}} & \edit{46.51} \\
 PSS & 33.33 & \edit{\textbf{74.67}} & \edit{39.88} \\
 POM & 16.67 & \edit{\textbf{44.39}} & \edit{23.23} \\
 \hline
 & \textbf{Random} & \textbf{\method-13b} & \edit{\textbf{\method-13b (no OPD)}} \\
 \hline
 PC & 33.33 & \edit{\textbf{55.06}} & \edit{40.70} \\
 PSS & 33.33 & \edit{\textbf{84.00}} & \edit{39.88} \\
 POM & 16.67 & \edit{\textbf{60.43}} & \edit{18.71} \\
 \hline
\end{tabular}
\label{table:main_results}
\end{table}

During tactile feature alignment and end-to-end fine-tuning, we trained \method{} with comparison tasks (i.e. PC, PSS and POM) to align its physical understanding of our physical properties and objects with our labels. We evaluated \method's physical understanding with the same single-step prompts used during training and on 500 question-answer pairs in total across the three tasks. The results for physical understanding of unseen test objects are shown in Table \ref{table:main_results}.

Our results show that \edit{both \method-7b and \method-13b perform well on all three physical understanding tasks when they are trained to predict property descriptions. Using physical property descriptions, \method-7b achieves accuracies of 48.10\% on PC, 74.67\% on PSS and 44.39\% on POM. \method-13b outperforms \method-7b by 6.96\% on PC, 9.33\% on PSS and 16.04\% on POM. This suggests that \method's physical understanding improves significantly with LLM size. 

Further, we explored the effect of using physical property descriptions by fine-tuning both \method-7b and \method-13b on the physical understanding tasks without intermediate physical property predictions. We found that predictions based on object properties notably improve physical understanding in both \method-7b and \method-13b.
}

\subsection{Scenario Reasoning}

\begin{table}
\centering
\caption{\textbf{Results on \dataset{} Scenario Reasoning Tasks.} \edit{During scenario reasoning, we do not provide ground-truth property descriptions. Our experiments show that leveraging object properties significantly improves scenario reasoning for \method.
}}
\begin{tabular}{l c c c}
 \hline
 & \textbf{Random} & \textbf{\method-7b} & \edit{\textbf{\method-13b}} \\
 \hline
 \edit{PSR} & 50.00 & \edit{\textbf{69.57}} &  \edit{67.39} \\
\edit{PSR (w/o OPD)} & 50.00 & \edit{\textbf{63.04}} & \edit{39.13} \\
 \hline
\end{tabular}
\label{table:psr_results}
\end{table}

We assessed the usefulness of our physical property categories by testing how \method{} can reason about everyday scenarios using the physical properties. For reference, the different scenario questions are provided in Table \ref{table:scenario} with \edit{the prompts shown in Table \ref{table:task_prompts}.

Our results are summarized in Table \ref{table:psr_results}. For both \method-7b and \method-13b, including the object property significantly improves performance, which supports our overall hypothesis that leveraging these properties is helpful for these tasks. Interestingly, we observed that the 7b model marginally outperformed the 13b model.
}

\begin{figure}
    \centering  \includegraphics[width=\linewidth]
    {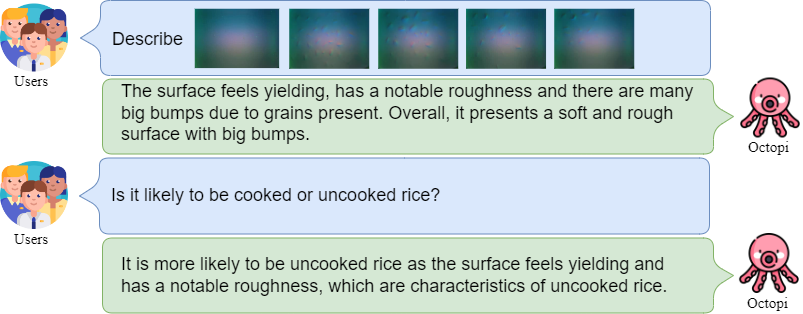}
    \caption{\textbf{Rice (Cooked v.s. Uncooked) Reasoning.} \edit{\method-13b is prompted to reason about whether a scoop of rice is more likely to be cooked or uncooked based on a tactile video of a scoop on uncooked rice. It reasons about the rice state correctly without being trained to do so.}}
    \label{fig:rice_state}
\end{figure}

\begin{figure}
    \centering
    \includegraphics[width=\linewidth]{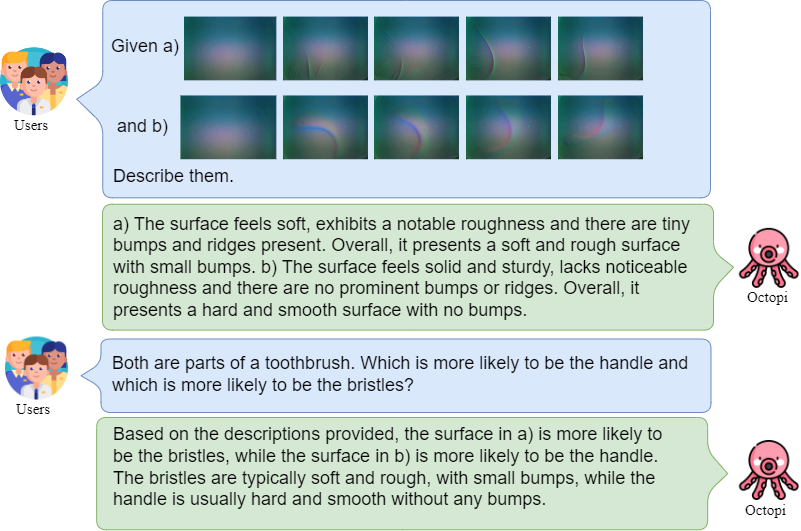}
    \caption{\textbf{Toothbrush Part Reasoning.} \edit{Given a tactile video of a toothbrush's handle and the same toothbrush's bristles, \method-13b is prompted to reason which tactile readings belong to the handle and which belongs to the bristles.}}
    \label{fig:object_part}
\end{figure}

We provide two qualitative examples to show \method-13b performing commonsense physical reasoning effectively. In the first task, we provide a tactile video of a scoop of uncooked rice and first instruct it to describe the tactile video. We then follow up with an instruction to \method-13b to determine if the rice is uncooked or cooked. \method-13b is able to reason that the scoop of rice is uncooked due to its rough surface, as shown in Fig. \ref{fig:rice_state}.

Next, we gave \method-13b two tactile videos corresponding to two different parts of the same toothbrush - the handle and the bristles. It is instructed to describe both objects using the physical properties. We then instruct it to determine which tactile video belongs to each object part using the physical properties. Fig. \ref{fig:object_part} shows that \method-13b is able to reason about the property-object match correctly.

\subsection{Avocado Ripeness Classification}

To evaluate \method's usefulness as a tactile-grounded physical reasoning system for real world tasks, we integrated two GelSight sensors on a 7-DoF Franka Emika Panda robot and used it for avocado ripeness classification. While ripe avocados generally appear in a shade of brown, their ripeness is difficult to determine using vision alone. At the same time, ripe avacados are softer then unripe ones and thus, tactile sensations can improve classification.

\edit{We performed property prediction and ripeness classification evaluations using a set of 10 avocados with 20 tactile samples collected from each avocado (i.e. 200 total samples). During ripeness classification, 100 pairs of avocado samples were selected and \method{} was tasked to identify which avocado is riper. Each pair contains avocados in different stages of ripeness.} At test time, the Franka robot grasped each avocado once to collect the tactile readings, before passing it to the model. The model gave instructions on which avocado to place in the ripe bin, and this was relayed through ROS for execution.

\begin{table}
\centering
\caption{\textbf{Avocado Property Prediction and Ripeness Classification Results.} \edit{\method-13b predicts avocado properties reasonably well with only a pressing motion. For avocado ripeness classification, \method-13b is able to leverage its commonsense knowledge to use both \textit{hardness} and \textit{bumpiness} properties.}}
\begin{tabular}{l c c c}
 \hline
 & \textbf{Random} & \textbf{\method-13b} & \edit{\textbf{PG-InstructBLIP}} \\
 \hline
 Property Prediction & 3.70 & \edit{35.50} & 0.00 \\
\hspace{3mm} Hardness & 33.33 & \edit{57.50} & \edit{37.50}  \\
\hspace{3mm} Roughness & 33.33 & \edit{71.00} & \edit{3.00}   \\
\hspace{3mm} Bumpiness & 33.33 & \edit{64.00} & \edit{9.50} \\
\hline 
 Ripeness Classification & 50.00 & \edit{63.00} & - \\
 \hline
\end{tabular}
\label{table:robot_results}
\end{table}

Table \ref{table:robot_results} shows \method-13b's performance on both avocado property prediction and avocado ripeness classification. We split the evaluation into two parts and evaluate the avocado property prediction and ripeness classification separately.

For avocado property prediction, \method-13b achieves an accuracy of \edit{35.50\%}, which is significantly higher than that of the random baseline (3.70\%). Unlike the \dataset{} dataset, these tactile videos are collected with only pressing and without any rotation. This suggests that \method-13b's physical property prediction capability is robust to differences in tactile exploratory procedures.

\edit{\method-13b obtains a ripeness prediction accuracy of 63.00\%. To set up this task, we first elicit \method's commonsense reasoning for avocado ripeness using our properties by querying ``You will be given tactile descriptions that consist of three physical properties: hardness, roughness, bumpiness. Hardness categories in increasing hardness: soft, moderately hard, hard. Roughness categories in increasing roughness: smooth, slightly smooth, rough. Bumpiness categories in increasing bumpiness: no bumps, small bumps, big bumps. Which of these properties help to determine avocado ripeness? Rank them." \method-13b responds with ``The properties that help determine avocado ripeness are hardness and bumpiness. Roughness is not a reliable indicator of avocado ripeness. In terms of hardness, a ripe avocado will be moderately hard, while an unripe avocado will be hard. In terms of bumpiness, a ripe avocado will have small bumps, while an unripe avocado will have no bumps or small bumps." This exchange is provided as context to \method-13b during ripeness classification.}

\edit{For comparison, we evaluated a physically-grounded vision-language model, PG-InstructBLIP \cite{gao2023physically}, on property predictions of our avocados. PG-InstructBLIP was trained to infer a predetermined set of physical properties from \emph{visual} images of real objects in the EgoObjects dataset \cite{zhu2023egoobjects}.
Table \ref{table:robot_results} shows PG-InstructBLIP's performance on property prediction for our avocados was poor. Possible reasons for this are that (i) the definitions of the physical properties may not be well-aligned with \dataset, and/or (ii) the physical properties of avocados are not clearly apparent using only the visual modality. We could not coax the PG-InstructBLIP model to directly classify avocado ripeness despite trying various prompts; it would always pick the first object.}

\subsection{Object Property Description Prediction}
\label{sec:unseen_results}

\begin{table}
\centering
\caption{\textbf{Results on \dataset{} Object Property Description Test Set.} \edit{FT CLIP is the combination of the fine-tuned CLIP visual encoder and the three separate trained classification layers. \method-7b and \method-13b perform above the random baseline for object property predictions and have similar performance to the fine-tuned CLIP. \method-13b performs better than \method-7b on the prediction task.}}
\begin{tabular}{l c c c c}
 \hline
 & \textbf{Random} & \textbf{FT CLIP} & \textbf{\method-7b} & \textbf{\method-13b} \\
 \hline
 Combined & 3.70 & \textbf{57.89} & \edit{47.37} & \edit{55.26} \\
 Hardness & 33.33 & \textbf{86.84} & \edit{71.05} & \edit{73.68} \\
 Roughness & 33.33 & 76.32 & \edit{73.68} & \edit{\textbf{78.95}} \\
 Bumpiness & 33.33 & 71.05 & \edit{\textbf{81.58}} & \edit{78.95} \\
 \hline
\end{tabular}
\label{table:main_results_opd}
\end{table}

The physical understanding and scenario reasoning capabilities of \method{} depends on its initial physical property predictions. We evaluated \method{}'s physical property prediction on the \dataset{} test set and show the results in Table \ref{table:main_results_opd}. Both \method-7b and \method-13b perform well above the random baseline for combined and individual property prediction and have similar performance to the fine-tuned CLIP model, indicating that \method{} can be  used for object property prediction. \edit{\method-13b has a higher combined accuracy (i.e. all three physical properties are correctly predicted for a given object) when compared to \method-7b, suggesting there are performance gains with larger LLMs for tactile signal grounding (apart from the \textit{bumpiness} property).}

\section{Ablations}
\label{sec:ablations}

In this section, we describe ablation studies to examine (i) the impact of the encoder's learned representations on physical property prediction and (ii) the influence of end-to-end fine-tuning data quantity on physical reasoning. For the following sections, we report test accuracy on unseen objects.

\subsection{Ablation: The Impact of Encoder Fine-tuning}

\begin{table}
\centering
\caption{\textbf{CLIP Fine-tuning Ablation Results on Object Property Prediction.} \edit{FT refers to fine-tuned. Using the CLIP fine-tuned on property prediction improves \method's performance in property prediction.}}
\begin{tabular}{l c c}
 \hline
 & \textbf{\method-7b (FT CLIP)} & \textbf{\method-7b (base CLIP)} \\
 \hline
 Combined & \edit{\textbf{47.37}} & \edit{39.47} \\
 Hardness & \edit{71.05} & \edit{\textbf{81.58}} \\
 Roughness & \edit{\textbf{73.68}} & \edit{52.63} \\
 Bumpiness & \edit{\textbf{81.58}} & \edit{55.26} \\
 \hline
 & \edit{\textbf{\method-13b (FT CLIP)}} & \edit{\textbf{\method-13b (base CLIP)}} \\
 \hline
 Combined & \edit{\textbf{55.26}} & \edit{50.00} \\
 Hardness & \edit{73.68} & \edit{\textbf{76.32}} \\
 Roughness & \edit{\textbf{78.95}} & \edit{65.79} \\
 Bumpiness & \edit{\textbf{78.95}} & \edit{76.32} \\
 \hline
\end{tabular}
\label{table:encoder_opd}
\end{table}

We used vision-based tactile inputs in this work and pre-trained vision foundation models (i.e. CLIP) have shown impressive performance on vision tasks. To test whether additional fine-tuning improves the pre-trained CLIP encoder's representations for physical property prediction using tactile images, we conducted ablation experiments. We compared the performance of two \method{} versions --- one trained with the off-the-shelf CLIP encoder and the other trained with the fine-tuned CLIP encoder. 

In Table \ref{table:encoder_opd}, our \textit{Object Property Description} results show that  \method-7b trained with a fine-tuned CLIP encoder outperforms  one trained with an unmodified CLIP encoder by 7.90\% on combined accuracy. Similarly, \method-13b with the fine-tuned CLIP visual encoder performs better on the combined, roughness, and bumpiness predictions, with the combined accuracy being 5.26\% higher. This suggests that a fine-tuned CLIP generally improves its learned representations for physical property prediction in an end-to-end LVLM.

\begin{table}
\centering
\caption{\textbf{CLIP Fine-tuning Ablation Results on Physical Understanding Tasks.} \edit{Using a fine-tuned CLIP improves \method's performance in physical understanding tasks for both \method-7b and \method-13b.}}
\begin{tabular}{l c c}
 \hline
 & \textbf{\method-7b (fine-tuned CLIP)} & \textbf{\method-7b (base CLIP)} \\
 \hline
 PC & \edit{\textbf{48.10}} & \edit{30.38} \\
 PSS & \edit{\textbf{74.67}} & \edit{42.67} \\
 POM & \edit{\textbf{44.39}} & \edit{36.36} \\
 \edit{PSR} & \edit{\textbf{69.57}} & \textbf{\edit{69.57}} \\
 \hline
 & \edit{\textbf{\method-13b (fine-tuned CLIP)}} & \edit{\textbf{\method-13b (base CLIP)}} \\
 \hline
 PC & \edit{\textbf{55.06}} & \edit{47.47} \\
 PSS & \edit{\textbf{84.00}} & \edit{75.33} \\
 POM & \edit{\textbf{60.43}} & \edit{57.22} \\
 \edit{PSR} & \edit{\textbf{67.39}} & \edit{45.65} \\
 \hline
\end{tabular}
\label{table:encoder_understanding}
\end{table}

\edit{We further tested both \method{} versions on physical understanding tasks with results in Table \ref{table:encoder_understanding}. For \method-7b, the version trained with a fine-tuned CLIP encoder performs better across the three physical understanding tasks (by 17.72\% on PC, 32.00\% on PSS, 8.03\% on POM). Similarly, \method-13b with the fine-tuned CLIP encoder has a better performance for physical understanding tasks, which suggests that fine-tuning generally helps physical understanding and physical reasoning performance.} Further encoder analysis can be found in Appendix \ref{app:encoders}.

\subsection{Ablation: The Impact of End-to-end Fine-tuning}

\begin{table}
\centering
\caption{\edit{\textbf{End-to-end Fine-tuning Physical Property Prediction Result Comparisons.} End-to-end fine-tuning with LoRA generally improves physical property prediction accuracies.}}
\begin{tabular}{l c c}
 \hline
 & \edit{\textbf{\method-7b (w/ LoRA)}} & \edit{\textbf{\method-7b (w/o LoRA)}} \\
 \hline
 Combined & \edit{\textbf{47.37}} & \edit{39.47} \\
 Hardness & \edit{\textbf{71.05}} & \edit{65.79} \\
 Roughness & \edit{73.68} & \edit{\textbf{76.32}} \\
 Bumpiness & \edit{\textbf{81.58}} & \edit{71.05} \\
 \hline
 & \edit{\textbf{\method-13b (w/ LoRA)}} & \edit{\textbf{\method-13b (w/o LoRA)}} \\
 \hline
 Combined & \edit{\textbf{55.26}} & \edit{23.68} \\
 Hardness & \edit{\textbf{73.68}} & \edit{36.84} \\
 Roughness & \edit{\textbf{78.95}} & \edit{73.68} \\
 Bumpiness & \edit{\textbf{78.95}} & \edit{71.05} \\
 \hline
\end{tabular}
\label{table:lora_properties}
\end{table}

\edit{Table \ref{table:lora_properties} shows \method{}'s performance on the property prediction task before and after end-to-end fine-tuning with LoRA. For both \method-7b and \method-13b, the fine-tuned variants generally performed better. We see sharp improvements for \method-13b with improvements across the properties. Our results suggest that end-to-end fine-tuning improves physical property prediction accuracy.}
\edit{Similar to the property prediction task, we observed that fine-tuning with LoRA also improves \method's performance on physical understanding tasks (Table \ref{table:lora_physical_understanding}).}

\begin{table}
\centering
\caption{\edit{\textbf{End-to-end Fine-tuning Physical Understanding Result Comparisons.} End-to-end fine-tuning for physical understanding tasks significantly improves physical understanding for both \method-7b and \method-13b.}}
\begin{tabular}{l c c c}
 \hline
 & \textbf{Rand.} & \edit{\textbf{\method-7b (LoRA)}} & \edit{\textbf{\method-7b (w/o LoRA)}} \\
 \hline
 PC & 33.33 & \edit{\textbf{48.10}} & \edit{40.51} \\
 PSS & 33.33 & \edit{\textbf{74.67}} & \edit{30.00} \\
 POM & 16.67 & \edit{\textbf{44.39}} & \edit{12.30} \\
 \edit{PSR} & 50.00 & \edit{\textbf{69.57}} & \edit{63.04} \\
 \hline
 & \textbf{Rand.} & \edit{\textbf{\method-13b (LoRA)}} & \edit{\textbf{\method-13b (w/o LoRA)}} \\
 \hline
 PC & 33.33 & \edit{\textbf{55.06}} & \edit{52.53} \\
 PSS & 33.33 & \edit{\textbf{84.00}} & \edit{77.33} \\
 POM & 16.67 & \edit{\textbf{60.43}} & \edit{34.76} \\
 \edit{PSR} & 50.00 & \edit{\textbf{67.39}} & \edit{60.87} \\
 \hline
\end{tabular}
\label{table:lora_physical_understanding}
\end{table}

\section{Conclusion and Discussion}
\label{sec:conclusion}
In this work, we extended large vision-language models (LVLMs) to process and describe tactile inputs using physical properties. \edit{We proposed a tactile dataset called \dataset{}, comprising  data from vision (Camera) and tactile (GelSight) sensors collected from everyday objects, along with physical property annotations. We also present \method, a large tactile-language model trained using datasets like \dataset{} to perform physical property reasoning using tactile inputs.}

Our experiments show that \method{} is able to describe tactile signals from novel unseen objects and that inferred physical properties can be  used for physical reasoning and robot task completion in scenarios with visual ambiguity. \edit{We studied the impact of different components in \method, and found that using a task-specific visual encoder that is fine-tuned on our labels improves performance significantly across all tasks. This suggests that improvements to the visual encoder will yield benefits. In addition, parameter-efficient LLM fine-tuning consistently improved performance.

Our work opens up future work in tactile robotics. We are currently working on tactile encoder improvement and more diverse exploratory procedures to obtain additional physical properties. It would also be interesting to combine different datasets (e.g., those using other tactile sensors~\cite{taunyazov2020event,taunyazov2021extended}), along with other modalities such as robot proprioception. We plan to also perform physical understanding alignment with object images and LLM fine-tuning with additional physical understanding data~\cite{wang2023newton, li-etal-2023-language-models}.
}

\section*{Acknowledgements}
This research is supported by the National Research Foundation, Singapore under its Medium Sized Center for Advanced Robotics Technology Innovation.

\balance

\bibliographystyle{plainnat}
\bibliography{references}

\clearpage 

\nobalance

\begin{strip}

\begin{center}
{\Huge 
Appendix for Octopi: Object Property Reasoning\\with Large Tactile-Language Models}
\end{center}
\end{strip}

In the appendices below, we provide more details for our \dataset{} dataset, specifically for the annotations, objects, properties, sample videos and prompts. We also provide details on the \method{} encoder and avocado experiments. 

\appendices

\setcounter{table}{0}
\renewcommand{\thetable}{A\arabic{table}}

\section{Annotation Details}

We developed a general set of guidelines to align the annotations for the properties and reduce  variance. We show the guidelines in Table \ref{table:annotation_guidelines}. While the \textit{hardness} and \textit{roughness} properties are generally annotated with actual haptic feedback from human exploratory procedures, the \textit{bumpiness} property is generally annotated visually from the GelSight tactile images.

\begin{table}[H]
\centering
\caption{\textbf{\dataset{} Property Category Annotation Guidelines.} \edit{For each property and its three categories, we detail the general guideline we used to choose the most appropriate category for a tactile sample. The annotator rating for each property and its categories are given as [X] where X is the rating.}}
\begin{tabular}{l l} 
 \hline
 \textbf{Property Category} & \textbf{Annotation Guideline} \\
 \hline
 Soft [0] & compressible upon pressing with little force \\
 Moderately hard [1] & compressible upon pressing with moderate force  \\
 Hard [2] & incompressible with high human pressing force \\
 \hline
 Smooth [0] & minimal feeling of friction upon finger sliding \\
 Slightly rough [1] & slight feeling of friction upon finger sliding \\
 Rough [2] & significant feeling of friction upon finger sliding \\
 \hline
 No bumps [0] & no bumps visible on object surface \\
 Small bumps [1] & bumps are less than $1/4$ of the tactile image \\
 Big bumps [2] & bumps are more than $1/4$ of the tactile image \\
 \hline
\end{tabular}
\label{table:annotation_guidelines}
\end{table}

\begin{table}[H]
\centering
\caption{\edit{\textbf{\dataset{} Annotation Ratings.} We show the annotation ratings for each property and its three categories for all objects. The ratings are given as [X, Y, Z] where X is the rating by annotator 1, Y by annotator 2 and Z by annotator 3. The mapping between the ratings and the property categories can be found in Table \ref{table:annotation_guidelines}.}}
\begin{tabular}{l c c c} 
 \hline
 \textbf{Sample Name} & \textbf{Hardness} & \textbf{Roughness} & \textbf{Bumpiness} \\
 \hline
 baseball & [2, 2, 2] & [2, 2, 2] & [0, 1, 0] \\
 basket & [2, 2, 2] & [2, 2, 2] & [2, 1, 2] \\
 blanket & [1, 1, 1] & [0, 0, 0] & [0, 0, 0] \\
 bread\_knife\_blade & [2, 2, 2] & [0, 0, 0] & [0, 0, 0] \\
 bread\_knife\_handle & [2, 2, 2] & [1, 2, 2] & [0, 0, 0] \\
 rubber\_bands & [1, 0, 2] & [1, 0, 1] & [2, 2, 2] \\
 wires & [2, 1, 2] & [0, 0, 0] & [2, 1, 2] \\
 cotton\_ball & [0, 0, 0] & [1, 0, 1] & [0, 0, 0] \\
 rug & [1, 0, 1] & [2, 1, 2] & [2, 2, 2] \\
 feather\_duster & [0, 0, 1] & [0, 0, 0] & [1, 1, 1] \\
 water\_bottle & [1, 1, 1] & [0, 0, 0] & [2, 2, 2] \\
 strainer\_handle & [2, 2, 2] & [0, 0, 0] & [0, 0, 0] \\
  \hline
\end{tabular}
\label{table:annotation_ratings}
\end{table}

\begin{table}[H]
\centering
 \begin{tabular}{l c c c} 
 \hline
 strainer\_base & [1, 2, 1] & [2, 2, 2] & [1, 2, 1] \\
 controller\_base & [2, 2, 2] & [0, 0, 0] & [0, 0, 0] \\
 controller\_buttons & [2, 2, 2] & [1, 1, 1] & [0, 1, 0] \\
 controller\_keypad & [2, 2, 2] & [1, 1, 1] & [2, 1, 2] \\
 controller\_shoulderpads & [2, 2, 2] & [0, 0, 1] & [1, 0, 1] \\
 controller\_stick & [2, 2, 2] & [1, 1, 1] & [1, 1, 1] \\
 gauze\_pad & [1, 2, 2] & [1, 1, 1] & [1, 2, 0] \\
 lanyard\_string & [1, 1, 2] & [0, 0, 1] & [1, 2, 0] \\
 lanyard\_card & [2, 2, 2] & [0, 0, 0] & [2, 0, 2] \\
 leather\_book & [2, 2, 2] & [0, 0, 0] & [0, 0, 0] \\
 lemon & [1, 1, 2] & [1, 1, 1] & [1, 2, 1] \\
 mandarin\_orange & [1, 1, 1] & [1, 1, 1] & [1, 1, 1] \\
 mop\_head & [1, 1, 1] & [1, 1, 1] & [0, 1, 1] \\
 rubber\_slippers & [1, 1, 1] & [1, 1, 1] & [2, 1, 1] \\
 scissor\_blade & [2, 2, 2] & [0, 0, 0] & [0, 0, 0] \\
 scissor\_handle & [2, 2, 2] & [0, 0, 0] & [0, 0, 0] \\
 paper\_towel & [0, 0, 0] & [1, 1, 1] & [1, 1, 2] \\
 avocado\_kinda\_ripe & [1, 2, 1] & [2, 2, 2] & [1, 1, 2] \\
 kiwi\_kinda\_ripe & [1, 1, 1] & [2, 2, 2] & [1, 2, 0] \\
 pen\_base & [2, 2, 2] & [0, 0, 0] & [0, 0, 0] \\
 pen\_pad & [1, 0, 2] & [1, 0, 1] & [2, 0, 2] \\
 pillow & [0, 0, 0] & [0, 0, 0] & [0, 0, 0] \\
 potato & [2, 2, 2] & [1, 1, 1] & [1, 1, 1] \\
 rice\_spatula\_handle & [2, 2, 2] & [0, 0, 0] & [0, 0, 0] \\
 rice\_spatula\_base & [2, 2, 2] & [2, 2, 2] & [2, 2, 2] \\
 masking\_tape & [1, 2, 2] & [1, 1, 1] & [1, 1, 0] \\
 toilet\_paper & [1, 1, 1] & [1, 1, 1] & [1, 2, 1] \\
 wood\_ruler & [2, 2, 2] & [0, 0, 0] & [0, 0, 0] \\
 rice & [1, 1, 2] & [2, 2, 2] & [2, 2, 2] \\
 aluminum\_foil & [1, 1, 2] & [1, 0, 1] & [1, 1, 2] \\
 stress\_ball & [0, 0, 1] & [0, 0, 0] & [0, 0, 0] \\
 tennis\_ball & [1, 2, 1] & [1, 1, 1] & [1, 1, 1] \\
 toilet\_brush\_handle & [1, 0, 2] & [0, 0, 0] & [0, 0, 0] \\
 tomato & [1, 1, 2] & [0, 0, 0] & [0, 0, 0] \\
 toothbrush\_handle & [2, 2, 2] & [0, 0, 0] & [2, 0, 2] \\
 tv\_remote\_back & [2, 2, 2] & [0, 0, 0] & [0, 0, 0] \\
 tv\_remote\_buttons & [1, 1, 2] & [2, 2, 2] & [1, 2, 0] \\
 eraser & [1, 1, 2] & [0, 0, 0] & [0, 0, 0] \\
 insulating\_holder & [0, 0, 1] & [2, 2, 2] & [1, 2, 1] \\
 orange & [1, 1, 1] & [1, 1, 1] & [1, 2, 1] \\
 oven\_mitt & [1, 1, 2] & [2, 2, 2] & [1, 2, 1] \\
 egg & [2, 2, 2] & [0, 0, 0] & [0, 0, 0] \\
 avocado\_unripe & [2, 2, 2] & [2, 2, 2] & [1, 2, 0] \\
 tissue\_ball & [0, 0, 0] & [0, 0, 0] & [1, 2, 0] \\
 hairbrush\_bristles & [2, 2, 2] & [2, 2, 2] & [2, 2, 1] \\
 sponge\_rough & [1, 1, 2] & [2, 2, 2] & [1, 2, 1] \\
 hairbrush\_bristles\_side & [1, 2, 1] & [2, 2, 2] & [2, 2, 2] \\
 toothbrush\_bristles & [0, 0, 1] & [2, 2, 2] & [1, 2, 1] \\
 sponge\_soft & [0, 0, 0] & [1, 1, 1] & [1, 2, 0] \\
 steel\_wool & [0, 0, 1] & [2, 2, 2] & [1, 1, 2] \\
 feather\_duster\_handle & [2, 2, 1] & [0, 0, 0] & [0, 0, 1] \\
 nylon\_shirt & [1, 1, 1] & [1, 1, 1] & [0, 2, 0] \\
 denim & [1, 1, 1] & [1, 1, 1] & [1, 2, 0] \\
 millet & [1, 1, 2] & [2, 2, 2] & [1, 2, 1] \\
 bubble\_wrap & [1, 1, 1] & [2, 1, 2] & [2, 1, 2] \\
 tsa\_lock\_numbers & [2, 2, 2] & [2, 2, 2] & [1, 1, 2] \\
 bath\_towel & [1, 1, 1] & [2, 2, 2] & [1, 2, 1] \\
 ice\_block & [2, 2, 2] & [0, 0, 0] & [0, 0, 0] \\
 clothes\_peg & [2, 2, 2] & [1, 0, 1] & [0, 1, 0] \\
 hairbrush\_handle & [2, 2, 2] & [0, 0, 0] & [0, 0, 0] \\
 microfiber\_cloth & [1, 1, 1] & [1, 1, 1] & [1, 1, 0] \\
 toilet\_brush\_bristles & [0, 0, 1] & [2, 2, 2] & [1, 2, 1] \\
 hairbrush\_bristles & [2, 2, 2] & [2, 2, 2] & [2, 2, 1] \\
 \hline
\end{tabular}
\label{table:annotation_ratings}
\end{table}

\clearpage 
\section{Object Details}
In Tables \ref{table:objects_training} and \ref{table:objects_val_test}, we provide the list of objects in \dataset, their associated video name, and the set (i.e. train/val/test). We use the object names in the left column during the \textit{Property-object Matching} task.

\begin{table}[H]
\centering
\caption{\textbf{\dataset{} Training Set Objects.} \edit{We list all objects in our training set and their sample names.}}
\begin{tabular}{l l l} 
 \hline
 \textbf{Object} & \textbf{Sample Name}  & \textbf{Split} \\
 \hline
 a baseball & baseball & train \\
 a basket & basket & train \\
 a blanket & blanket & train \\
 a bread knife's blade & bread\_knife\_blade & train \\
 a bread knife's handle & bread\_knife\_handle & train \\
 a bunch of rubber bands & rubber\_bands & train \\
 a bunch of wires & wires & train \\
 a cotton ball & cotton\_ball & train \\
 a dishwashing cloth & rug & train \\
 a feather duster's head & feather\_duster & train \\
 a filled disposable water bottle & water\_bottle & train \\
 a fine mesh strainer's handle & strainer\_handle & train \\
 a fine mesh strainer's scoop & strainer\_base & train \\
 a game controller's body & controller\_base & train \\
 a game controller's buttons & controller\_buttons & train \\
 a game controller's keypad & controller\_keypad & train \\
 a game controller's shoulder buttons & controller\_shoulderpads & train \\
 a game controller's thumbstick & controller\_stick & train \\
 a gauze pad & gauze\_pad & train \\
 a lanyard & lanyard\_string & train \\
 a lanyard's card holder & lanyard\_card & train \\
 a leather book cover & leather\_book & train \\
 a lemon & lemon & train \\
 a mandarin orange & mandarin\_orange & train \\
 a mop's head & mop\_head & train \\
 a pair of rubber slippers & rubber\_slippers & train \\
 a pair of scissors' blade & scissor\_blade & train \\
 a pair of scissors' handle & scissor\_handle & train \\
 a sheet of paper towel & paper\_towel & train \\
 a ripe avocado & avocado\_kinda\_ripe & train \\
 a ripe kiwi & kiwi\_kinda\_ripe & train \\
 a pen's barrel & pen\_base & train \\
 a pen's grip & pen\_pad & train \\
 a pillow & pillow & train \\
 a potato & potato & train \\
 a rice spatula's handle & rice\_spatula\_handle & train \\
 a rice spatula's scoop & rice\_spatula\_base & train \\
 a roll of masking tape & masking\_tape & train \\
 a roll of toilet paper & toilet\_paper & train \\
 a ruler & wood\_ruler & train \\
 a scoop of rice & rice & train \\
 a sheet of aluminium foil & aluminium\_foil & train \\

 a stress ball & stress\_ball & train \\
 a tennis ball & tennis\_ball & train \\
 a toilet brush's handle & toilet\_brush\_handle & train \\
 a tomato & tomato & train \\
 a toothbrush's handle & toothbrush\_handle & train \\
 a TV remote's back & tv\_remote\_back & train \\
 a TV remote's buttons & tv\_remote\_buttons & train \\
 an eraser & eraser & train \\
 an insulating holder & insulating\_holder & train \\
 an orange & orange & train \\
 an oven mitt & oven\_mitt & train \\
 an uncracked egg & egg & train \\
 an unripe avocado & avocado\_unripe & train \\
 crumpled tissue paper & tissue\_ball & train \\
 the ends of a hairbrush's bristles & hairbrush\_bristles & train \\
 the rough side of a sponge & sponge\_rough & train \\
 the sides of a hairbrush's bristles & hairbrush\_bristles\_side & train \\
 the sides of a toothbrush's bristles & toothbrush\_bristles & train \\
 the soft side of a sponge & sponge\_soft & train \\
 \hline
\end{tabular}
\label{table:objects_training}
\end{table}

\begin{table}
\centering
\caption{\textbf{\dataset{} Validation and Test Set Objects.} \edit{We list all objects in our validation and test sets and their sample names.}}
\begin{tabular}{l l l}
 \hline
 \textbf{Object} & \textbf{Sample Name}  & \textbf{Split} \\
 \hline
 a ball of steel wool & steel\_wool & val \\
 a feather duster's handle & feather\_duster\_handle & val \\
 a nylon shirt & nylon\_shirt & val \\
 a pair of jeans & denim & val \\
 a scoop of millet & millet & val \\
 a sheet of bubble wrap & bubble\_wrap & val \\
 numbers on a TSA lock & tsa\_lock\_numbers & val \\
 \hline
a bath towel & bath\_towel & test \\
 a block of ice & ice\_block & test \\
 a clothes peg & clothes\_peg & test \\
 a hairbrush's handle & hairbrush\_handle & test \\
 a microfiber cloth & microfiber\_cloth & test \\
 a toilet brush's bristles & toilet\_brush\_bristles & test \\
 the ends of a hairbrush's bristles & hairbrush\_bristles & test \\
 \hline
\end{tabular}
\label{table:objects_val_test}
\end{table}

In Table \ref{table:object_parts}, we list objects that have semantically meaningful parts for part-based physical reasoning tasks. Specifically, we note that their semantic parts can be used for more precise \textit{Property-object Matching}. Surface materials of \dataset{} objects are listed in Table \ref{table:materials}. \dataset{} contains nine different common materials: fabric, food, leather, metal, paper, plastic, rubber, silicone and wood. While we do not use these materials for any task currently, we catalog them for future extensions (e.g. material classification using properties).

\begin{table}[H]
\centering
\caption{\textbf{\dataset{} Object Parts.} \edit{We list the objects in \dataset{} that have distinct semantic components, the names of their semantic components and the components' associated sample names.}}
\begin{tabular}{l l l} 
 \hline
 \textbf{Object} & \textbf{Part} & \textbf{Part Sample Name} \\
 \hline
 Bread knife & Blade & bread\_knife\_blade \\
 & Handle & bread\_knife\_handle \\
 \hline
 Feather duster & Handle & feather\_duster\_handle \\
 & Head & feather\_duster \\
 \hline
 Fine mesh strainer & Handle & strainer\_handle \\
 & Scoop & strainer\_base \\
 Game controller & Body & controller\_base \\
 & Buttons & controller\_buttons \\
 & Keypad & controller\_keypad \\
 & Shoulder buttons & controller\_shoulderpads \\
 & Thumbstick & controller\_stick \\
 \hline
 Hairbrush & Ends of bristles & hairbrush\_bristles \\
 & Handle & hairbrush\_handle \\
 & Sides of bristles & hairbrush\_bristles\_side \\
 \hline
 Lanyard & Card holder & lanyard\_card \\
 & String & lanyard\_string \\
 \hline
 Pair of scissors & Blade & scissor\_blade \\
 & Handle & scissor\_handle \\
  \hline
 Pen & Barrel & pen\_base \\
 & Grip & pen\_pad \\
 \hline
 Rice spatula & Handle & rice\_spatula\_handle \\
 & Scoop & rice\_spatula\_base \\
 \hline
 Sponge & Rough side & sponge\_rough \\
 & Soft side & sponge\_soft \\
 \hline
 Toilet brush & Bristles & toilet\_brush\_bristles \\
 & Handle & toilet\_brush\_handle \\
 \hline
 Toothbrush & Handle & toothbrush\_handle \\
 & Sides of bristles & toothbrush\_bristles \\
 \hline
 TV remote & Back & tv\_remote\_back \\
 & Buttons & tv\_remote\_buttons \\
 \hline
\end{tabular}
\label{table:object_parts}
\end{table}

\begin{table}[H]
\centering
\caption{\textbf{\dataset{} Object Surface Materials} in \dataset{} and their associated objects.}
\begin{tabular}{l l} 
 \hline
 \textbf{Surface Material} & \textbf{Object} \\
 \hline
 fabric & bath\_towel \\
 & blanket \\
 & cotton\_ball \\
 & denim \\
 & feather\_duster \\
 & gauze\_pad \\
 & hairbrush\_bristles \\
 & hairbrush\_bristles\_side \\
 & lanyard\_string \\
 & microfiber\_cloth \\
 & mop\_head \\
 & nylon\_shirt \\
 & oven\_mitt \\
 & pillow \\
 & rug \\
 & tennis\_ball \\
 & toothbrush\_bristles \\
 \hline
 food & avocado\_kinda\_ripe \\
 & avocado\_unripe \\
 & egg \\
 & ice\_block \\
 & kiwi\_kinda\_ripe \\
 & lemon \\
 & mandarin\_orange \\
 & millet \\
 & orange \\
 & potato \\
 & rice \\
 & tomato \\
 \hline
 leather & baseball \\
 & leather\_book \\
 \hline
 metal & aluminium\_foil \\
 & bread\_knife\_blade \\
 & scissor\_blade \\

 & steel\_wool \\
 & strainer\_handle \\
 & strainer\_base \\
 & tsa\_lock\_numbers \\
 \hline
 paper & masking\_tape \\
 & paper\_towel \\
 & tissue\_ball \\
 & toilet\_paper \\
 \hline
 plastic & bubble\_wrap \\
 & clothes\_peg \\
 & controller\_base \\
 & controller\_buttons \\
 & controller\_keypad \\ 
 & controller\_shoulderpads \\
 & controller\_stick \\
 & feather\_duster\_handle \\
 & hairbrush\_handle \\
 & lanyard\_card \\
 & pen\_base \\
 & rice\_spatula\_base \\
 & rice\_spatula\_handle \\
 & scissor\_handle \\
 & sponge\_rough \\
 & sponge\_soft \\
 & toilet\_brush\_bristles \\
 & toilet\_brush\_handle \\
 & toothbrush\_handle \\
 & tv\_remote\_back \\
 & water\_bottle \\
 & wires \\
 \hline
 rubber & eraser \\
 & pen\_pad \\
 & rubber\_bands \\
 & rubber\_slippers \\
 & stress\_ball \\
\hline
 silicone & insulating\_holder \\
 & tv\_remote\_buttons \\
 \hline
 wood & basket \\
 & bread\_knife\_handle \\
 & wood\_ruler \\
 \hline
\end{tabular}
\label{table:materials}
\end{table}

\section{Property Statistics}

We show the object count and percentage of each property category in Table \ref{table:property_count}. There is at least 18.92\% of each category for each property such that the dataset is sufficiently balanced.

\begin{table}[H]
\centering
\caption{\textbf{\dataset{} Property Category Count.}}
\begin{tabular}{l c c} 
 \hline
 \textbf{Property Category} & \textbf{Object Count} & \textbf{Percentage (\%)} \\
 \hline
 Soft & 26 & 35.14 \\
 Moderately hard & 14 & 18.92 \\
 Hard & 34 & 45.95 \\
 \hline
 Smooth & 36 & 48.65 \\
 Slightly rough & 16 & 21.62 \\
 Rough & 22 & 29.73 \\
 \hline
 No bumps & 26 & 35.14 \\
 Small bumps & 34 & 45.95 \\
 Big bumps & 14 & 18.92 \\
 \hline
\end{tabular}
\label{table:property_count}
\end{table}

Table \ref{table:property_combo_count} shows the count for each combination of the three properties: \textit{hardness}, \textit{roughness} and \textit{bumpiness}. The joint distribution of these properties across everyday objects is not uniform. Some combinations are much more common (e.g. [\textit{Hard, smooth, no bumps}] for human-made objects) than others (e.g., [\textit{Soft, rough, no bumps}]). Furthermore, the curse of dimensionality means that it is difficult to represent all combinations well as the number of properties increases. Our future work will incorporate better representation learning methods to handle this imbalance.

\begin{table}[H]
\centering
\caption{\textbf{\dataset{} Property Combination Count.} \edit{We list all property combinations and the count for each combination.}}
\begin{tabular}{l c} 
 \hline
 \textbf{Property Combination} & \textbf{Count} \\
 \hline
 Soft, smooth, no bumps & 5 \\
 Soft, slightly rough, no bumps & 0 \\
 Soft, rough, no bumps & 0 \\
 \hline
 Soft, smooth, small bumps & 2 \\
 Soft, slightly rough, small bumps & 7 \\
 Soft, rough, small bumps & 8 \\
 \hline
 Soft, smooth, big bumps & 1 \\
 Soft, slightly rough, big bumps & 0 \\
 Soft, rough, big bumps & 3 \\
 \hline
 Moderately hard, smooth, no bumps & 2 \\
 Moderately hard, slightly rough, no bumps & 0 \\
 Moderately hard, rough, no bumps & 0 \\
 \hline
 Moderately hard, smooth, small bumps & 2 \\
 Moderately hard, slightly rough, small bumps & 4 \\
 Moderately hard, rough, small bumps & 3 \\
 \hline
 Moderately hard, smooth, big bumps & 2 \\
 Moderately hard, slightly rough, big bumps & 0 \\
 Moderately hard, rough, big bumps & 1 \\
 \hline
 Hard, smooth, no bumps & 17 \\
 Hard, slightly rough, no bumps & 1 \\
 Hard, rough, no bumps & 1 \\
 \hline
 Hard, smooth, small bumps & 1 \\
 Hard, slightly rough, small bumps & 4 \\
 Hard, rough, small bumps & 3 \\
 \hline
 Hard, smooth, big bumps & 4 \\
 Hard, slightly rough, big bumps & 0 \\
 Hard, rough, big bumps & 3 \\
 \hline
\end{tabular}
\label{table:property_combo_count}
\end{table}

\clearpage 

\section{Sample Video Statistics}

There are five to seven tactile videos for each object in \dataset. Table \ref{table:video_stats} shows the average, minimum and maximum number of frames in our tactile videos at 112.30, 50 and 126 respectively.

\begin{table}[H]
\centering
\caption{\textbf{Sample Video Length Statistics.} \edit{The average tactile sample video has 112.30 frames, a minimum of 50 frames and a maximum of 126 frames.}}
\begin{tabular}{c c c} 
 \hline
\textbf{Average} & \textbf{Min} & \textbf{Max} \\
 \hline
 112.30 & 50 & 126 \\
 \hline
\end{tabular}
\label{table:video_stats}
\end{table}

\subsection{Language Prompt Details}

We list the questions for each task in this section.

\begin{lstlisting}
object_property_description = [
    "Describe the physical properties of <tact_start> <img_tokens> <tact_end>.",
    "How does this tactile video <tact_start> <img_tokens> <tact_end> feel?"
]
    
property_comparison = [
    "Is the object in the tactile video <tact_start> <img_tokens> <tact_end> <more_property> than the one in <tact_start> <img_tokens> <tact_end>? Describe both objects before answering.",
    "Is the object in <tact_start> <img_tokens> <tact_end> <more_property> than the object in <tact_start> <img_tokens> <tact_end>? Describe both objects before answering.",
    "Is the object in the tactile video <tact_start> <img_tokens><tact_end> <less_property> than the one in <tact_start> <img_tokens> <tact_end>? Describe both objects before answering.",
    "Is the object in <tact_start> <img_tokens> <tact_end> <less_property> than the object in <tact_start> <img_tokens> <tact_end>? Describe both objects before answering."
]

property_superlative_selection = [
    "Given three tactile videos: a) <tact_start> <img_tokens> <tact_end>, b) <tact_start> <img_tokens> <tact_end>, c) <tact_start> <img_tokens> <tact_end>. Describe each object and then select the <most_property>.",
    "You have tactile videos of one object each: a) <tact_start> <img_tokens> <tact_end>, b) <tact_start> <img_tokens> <tact_end>, c) <tact_start> <img_tokens> <tact_end>. Describe each object and then select the <most_property> object.",
    "Given these tactile videos: a) <tact_start> <img_tokens> <tact_end>, b) <tact_start> <img_tokens> <tact_end>, c) <tact_start> <img_tokens> <tact_end>. Describe each object and then select the <least_property>.",
    "You have tactile videos of one object each: a) <tact_start> <img_tokens> <tact_end>, b) <tact_start> <img_tokens> <tact_end>, c) <tact_start> <img_tokens> <tact_end>. Describe each object and then select the <least_property> object."
]

property_object_match = [
    "Given three tactile videos: a) <tact_start> <img_tokens> <tact_end>, b) <tact_start> <img_tokens> <tact_end>, c) <tact_start <img_tokens <tact_end>. Describe the object in each video, then match each video to one of the following objects in alphabetical order: 1) <object>, 2) <object>, 3) <object>.",
    "You have tactile videos of one object each: a) <tact_start> <img_tokens> <tact_end>, b) <tact_start> <img_tokens> <tact_end>, c) <tact_start> <img_tokens> <tact_end>. Describe the object in each video, then match each video to one of the following objects in alphabetical order: 1) <object>, 2) <object>, 3) <object>."
]

property_scenario_reasoning = [
    "<scenario_question>. Select only one most appropriate object for this scenario based on physical property descriptions of the two objects. Use the format 'The most suitable object is x), because xxx'"
]

\end{lstlisting}

\begin{figure*}[!ht]
    \centering
\includegraphics[width=0.8\linewidth]{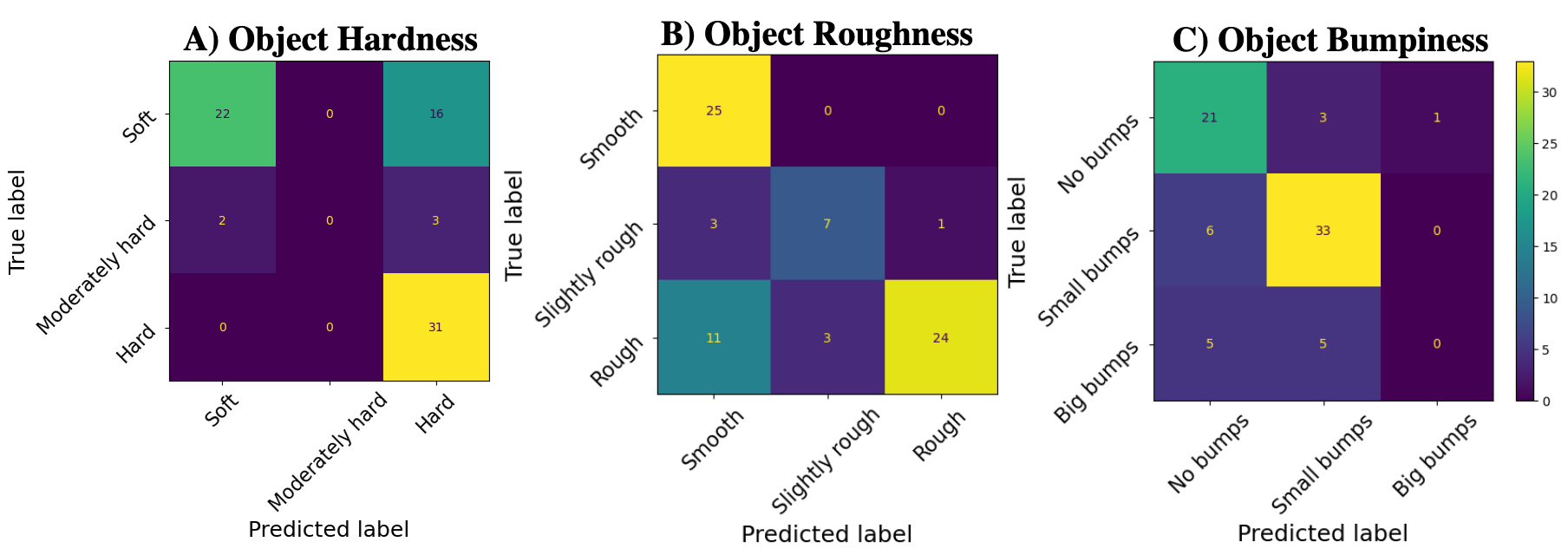}
    \caption{\textbf{Confusion Matrices for the CLIP Classifier's Physical Property Predictions.} We visualize the confusion matrices for the fine-tuned CLIP classifier's physical property predictions on unseen samples. 
    }
\label{fig:conf}
\end{figure*}

\begin{figure*}[!ht]
    \centering
    \includegraphics[width=0.8\linewidth]{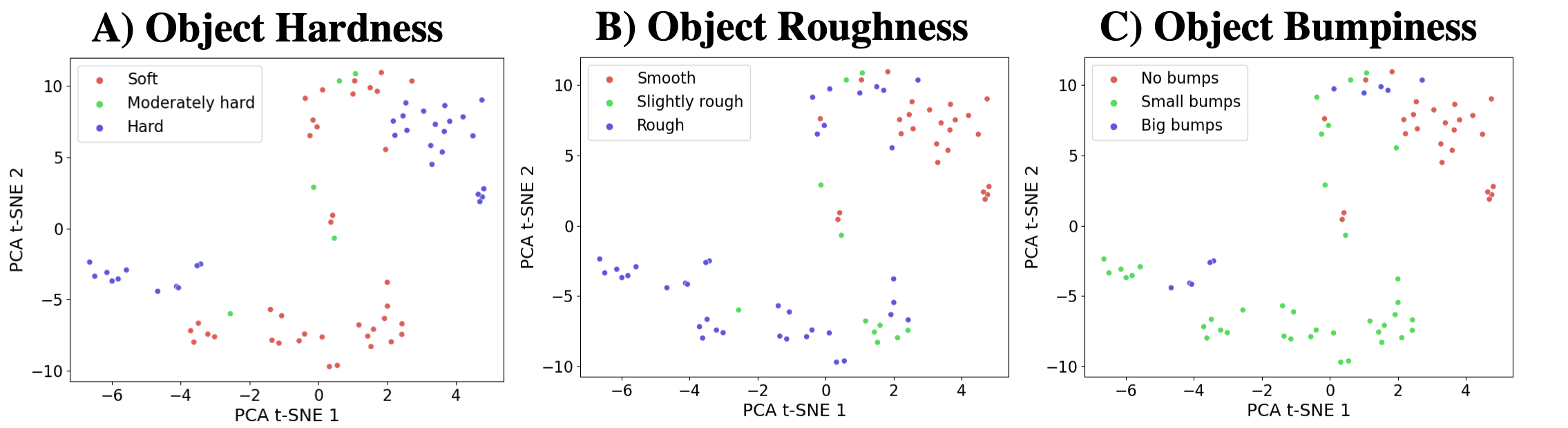}
    \caption{\textbf{Visualizations of CLIP Visual Encoder's Embeddings.} We visualize the fine-tuned CLIP visual encoder's output embeddings for each tactile video sample for each physical property. 
    }
\label{fig:tsne}
\end{figure*}

\section{Encoder Analysis}
\label{app:encoders}

Fig. \ref{fig:conf} visualizes the confusion matrices for the fine-tuned CLIP classifier's predictions and Fig. \ref{fig:tsne} plots its visual encoder's output embeddings for all unseen objects (dimensionality-reduced). 
We observe that the CLIP classifier performs well on \textit{Hard} objects but not as well on the \textit{Soft} and \textit{Moderately hard} objects. This is also evident in the encoder embeddings in the leftmost part of Fig. \ref{fig:tsne} where the embeddings for the \textit{Hard} objects are well-separated while the ones for \textit{Moderately hard} objects are especially not well-separated. For the roughness property, the CLIP classifier performs well on \textit{Smooth} and \textit{Slightly rough} objects but not as well on the \textit{Rough} objects. We can see this in the encoder embeddings in Fig. \ref{fig:tsne} where some of the embeddings for the \textit{Rough} objects are interspersed with the embeddings for the \textit{Smooth} objects. Lastly, for the bumpiness property, the CLIP is able to discriminate \textit{No bumps} and \textit{Small bumps} but not as well on objects with \textit{Big bumps}. In the encoder embeddings at the right side of Fig. \ref{fig:tsne}, the embeddings for the objects with \textit{Big bumps} are interspersed with the embeddings for the objects with \textit{No bumps} and \textit{Small bumps}.

\section{PG-InstructBLIP Avocado Property Prediction}

\begin{figure}
    \centering
\includegraphics[width=0.5\linewidth]{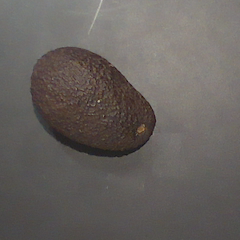}
    \caption{{Cropped Avocado Image for Vision-only Property Prediction.}}
    \label{fig:avocado_cropped}
\end{figure}

For the avocado experiments with PG-InstructBLIP, we take 20 RGB images of each avocado from a top-down view on a flat gray surface. We vary the position and orientation of each avocado for each image and use cropped images (example shown in Fig. \ref{fig:avocado_cropped}); the cropping of the avocado images makes it closer to the dataset PG-InstructBLIP was trained on. We use the original prompt provided in PG-InstructBLIP's example script and prompt the model for each physical property:
\begin{itemize}
    \item Hardness - ``Classify the object as hard, soft or medium? Respond unknown if you are not sure. Short answer:"
    \item Roughness - ``Classify the object as rough, smooth or medium? Respond unknown if you are not sure. Short answer:"
    \item Bumpiness - ``Classify the object as having big bumps, small bumps or no bumps? Respond unknown if you are not sure. Short answer:"
\end{itemize}

PG-InstructBLIP is not trained on our three physical properties and we find that it never chooses ``moderately hard" and ``slightly rough". Hence, we change those categories to ``medium" for hardness and roughness.

\end{document}